\documentclass{article} %
\usepackage{iclr2026_conference,times}

\usepackage{amsmath,amsfonts,bm}

\usepackage{hyperref}
\usepackage{colortbl}
\usepackage{url}
\usepackage{adjustbox}
\usepackage{booktabs}
\usepackage{multirow}
\usepackage{arydshln}
\usepackage{pifont}
\usepackage{xspace}
\usepackage{wrapfig}
\usepackage{placeins}
\usepackage{caption}
\usepackage{subcaption}
\usepackage[dvipsnames]{xcolor}
\usepackage{titlesec}
\usepackage{titletoc}
\usepackage{bbding}
\usepackage{amsmath}

\definecolor{blue_stitch}{HTML}{5088B8}
\definecolor{lightblue_stitch}{HTML}{6AC5E0}
\definecolor{purple_stitch}{HTML}{bd9ac4}
\definecolor{green_stitch}{HTML}{d9ead3}
\definecolor{blue_method}{HTML}{6bc5e0}

\newcommand{\phasetwo}[1]{\textcolor{black}{\textbf{#1}}}

\title{Stitch: Training-Free Position Control\\ in Multimodal Diffusion Transformers}

\author{
\hspace{-10px}
\begin{tabular}{c}
\vspace{5px}\\
Jessica Bader$^{1,2,3,*}$ \quad Mateusz Pach$^{1,2,3,*}$ \\
Mar\'ia A. Bravo$^{1,2,3}$ \quad
Serge Belongie$^{4}$ \quad Zeynep Akata$^{1,2,3}$ \\
\normalfont
$^1$Technical University of Munich \, 
$^2$Helmholtz Munich \, 
$^3$Munich Center for Machine Learning \\
\normalfont
$^4$University of Copenhagen \, 
$^*$equal contribution \, 
\texttt{jessica.bader@tum.de}
\end{tabular}
}

\newcommand{\cmark}{\ding{51}} %
\newcommand{\xmark}{\ding{55}} %
\newcommand{\methodname}{\textcolor{black}{\fontfamily{lmss}\selectfont Stitch}\xspace}

\newcommand{\methodnamebf}{\textcolor{black}{\fontfamily{lmss}\bfseries Stitch}\xspace}

\newcommand{\benchmarkname}{\mbox{PosEval}\xspace}

\iclrfinalcopy %
\begin{document}

\maketitle

\begin{abstract}
Text-to-Image (T2I) generation models have advanced rapidly in recent years, but accurately capturing spatial relationships like “above” or “to the right of” poses a persistent challenge. Earlier methods improved spatial relationship following with external position control. However, as architectures evolved to enhance image quality, these techniques became incompatible with modern models. We propose \methodname, a training-free method for incorporating external position control into Multi-Modal Diffusion Transformers (MMDiT) via automatically-generated bounding boxes. \methodname produces images that are both spatially accurate and visually appealing by generating individual objects within designated bounding boxes and seamlessly stitching them together. We find that targeted attention heads capture the information necessary to isolate and cut out individual objects mid-generation, without needing to fully complete the image. We evaluate \methodname on PosEval, our benchmark for position-based T2I generation. Featuring five new tasks that extend the concept of \textit{Position} beyond the basic GenEval task, PosEval demonstrates that even top models still have significant room for improvement in position-based generation. Tested on Qwen-Image, FLUX, and SD3.5, \methodname consistently enhances base models, even improving FLUX by $218\%$ on GenEval's \textit{Position} task and by $206\%$ on PosEval. \methodname achieves state-of-the-art results with Qwen-Image on PosEval, improving over previous models by $54\%$, all accomplished while integrating position control into leading models training-free. Code is available at~\url{https://github.com/ExplainableML/Stitch}.
\end{abstract}

\section{Introduction}
Text-to-Image (T2I) generation models provide a powerful bridge between natural language and visual content. By transforming textual descriptions into images, they expand human creativity, accelerate prototyping, and enable a wide range of applications. But as their popularity grows, so do user demands. These models should be able to generate real-world scenes with uncommon but still physically plausible object arrangements. They should interpret and fulfill requests that are both complex and linguistically nuanced in their descriptions of positioning. All of these needs could be met with strong position understanding, yet even basic spatial concepts such as “above” or “to the right of” remain a persistent challenge for current models~\citep{ghosh2023geneval}, and the difficulty only grows with more complex prompts.
Luckily, positional performance can be improved by augmenting off-the-shelf models with bounding boxes generated by Large Language Models (LLMs), but recent architectural upgrades have made traditional position-control methods incompatible with modern models. Nevertheless, these modifications have also enhanced both image quality and speed, traits that could be even more impactful if they were paired with precise positional prompt-following.

\begin{figure}[t]
    \centering
    \includegraphics[width=1.0\textwidth]{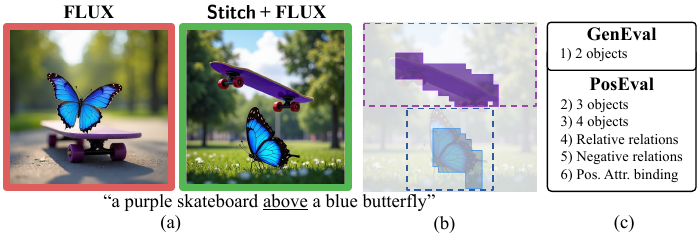} %
    \caption{(a) \textbf{\methodname} boosts position-aware generation, training-free, (b) by generating objects in LLM-made bounding boxes (dashed lines) and using attention heads for tighter latent segmentation mid-generation (filled). (c) Our \textbf{\benchmarkname} benchmark extends GenEval with 5 new positional tasks.}
    \label{fig:teaser}
\end{figure}

\begin{figure}[t]
    \centering
    \includegraphics[width=1.0\textwidth]{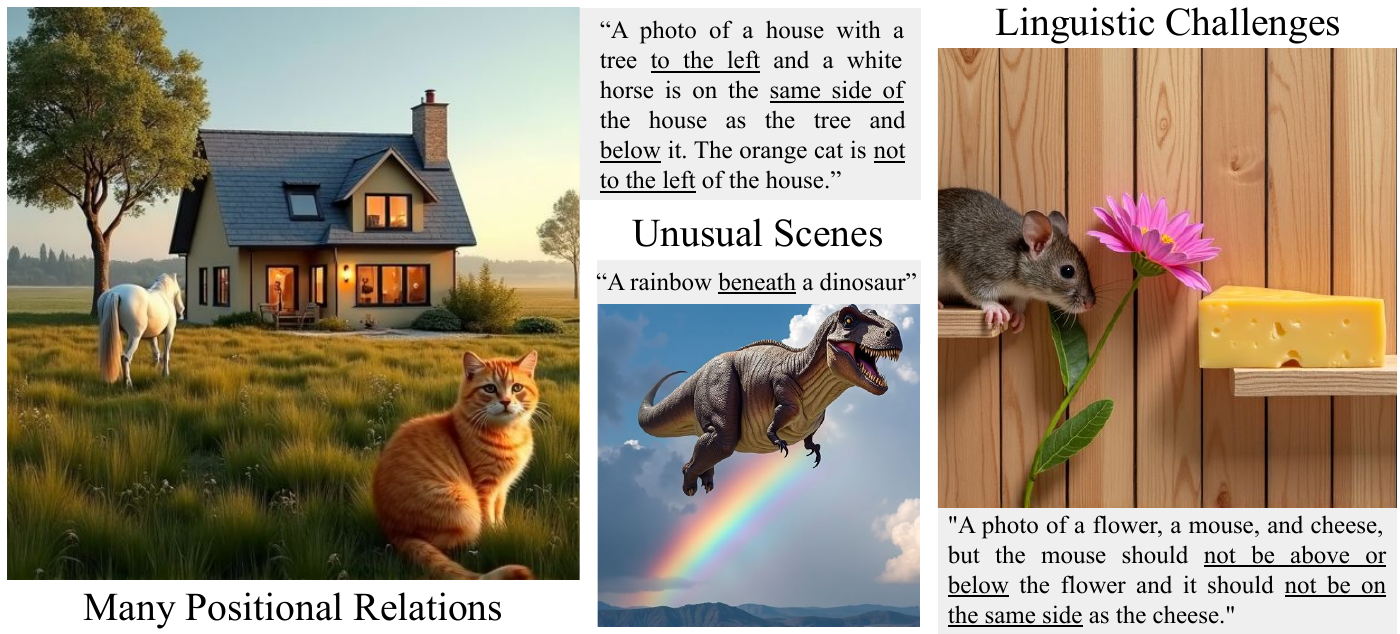} %
    \caption{\methodname excels at complex positional prompts.
    }
    \label{fig:qualitative2}
    \vspace{-10px}
\end{figure}

To bridge this gap, we propose \methodname: a test-time technique that effectively improves positional prompt generation (Figure~\ref{fig:teaser}a) by integrating additional position support via LLM-generated bounding boxes into leading Flow Matching (FM) T2I models based on Multi-Modal Diffusion Transformer (MMDiT) architecture. \methodname~achieves controlled generation by independently generating objects for $S$ steps, with each object constrained to its respective bounding box through attention modulation. Next, we extract foreground objects and combine the predictions. In particular, we find that foreground masks can be inexpensively derived directly from the attention heads (Figure~\ref{fig:teaser}b), on-the-fly before completing generation and without requiring an external model. After step $S$ all constraints are lifted, enabling the T2I model to refine the image organically in the remaining steps. Consequently, \methodname~facilitates quick and affordable upgrades in the positional performance of the leading T2I models, combining the best image quality with strong positional generation (Figure~\ref{fig:qualitative2}).

Such positional upgrades are extremely valuable for improving 2D spatial performance. Because while recent successes on \textit{Position} tasks in existing benchmarks such as GenEval~\citep{ghosh2023geneval} might suggest that the problem is nearing resolution, a closer analysis shows that it is far from solved. We take the next step in evaluating whether T2I models can consistently and robustly generate images that accurately reflect positional prompts. To this end, we introduce \benchmarkname, an extension of the GenEval~\citep{ghosh2023geneval} benchmark designed for in-depth evaluation of positional abilities in T2I generation, going beyond the traditional \textit{Position} category (Figure~\ref{fig:teaser}c). \benchmarkname includes five new tasks, each aimed at probing specific failure modes in T2I models. Using \benchmarkname, we demonstrate that state-of-the-art (SOTA) models continue to struggle with positional tasks, highlighting considerable room for improvement. \benchmarkname additionally provides a comprehensive evaluation platform that rigorously measures \methodname's effectiveness in improving positional generation, showcasing significant performance improvements compared to baseline models.

The primary contributions of this work are: (1) We introduce \methodname, a test-time method %
that substantially improves MMDiT-based models' capacity to accurately generate images from position-based prompts; (2) We find that certain attention heads encode sufficient information to extract the foreground object from the background within the latent space, long before the image is fully generated; and (3) We present the \benchmarkname benchmark, a GenEval~\citep{ghosh2023geneval} extension featuring five new targeted, position-focused tasks designed to tackle the next level of generation challenges.

\section{Related Works}
\paragraph{T2I generation.} T2I synthesis has rapidly advanced in the past years~\citep{Podell2023SDXLIL, dalle3-2023, midjourneyv7, gpt4o, imagen4}, moving from diffusion-based models like Stable Diffusion~\citep{rombach2021highresolution}, Imagen~\citep{Saharia2022ImagenPT}, and DALL·E~\citep{openai2023dalle3} to more powerful FM approaches like FLUX~\citep{flux2024}, SD3.5~\citep{esser2024scaling}, and HiDream~\citep{cai2025hidream}, advancing fidelity, diversity, and controllability. They enhance creativity, productivity, and communication~\citep{chen2023controlstyle, zhou2024creativityai, Zhou2024StoryDiffusionCS}. Previous works have improved pre-trained T2I models by enhancing prompt adherence~\citep{Eyring2024ReNOEO, Eyring2025NoiseHA}, alignment with human preferences~\citep{Karthik2024ScalableRP, richhf, bravo2025tialign, sendera2025semu}, and personalization~\citep{liu2023cones2, Ruiz2022DreamBoothFT, Kim2024DataDreamFG}. Despite these significant advances, positional control remains underexplored, and T2I models still struggle to manage object positions accurately~\citep{bakr2023hrs, Gokhale2022BenchmarkingSR}.

\paragraph{Position control improvements.} Many base models improve spatial limitations during end-to-end training, for example with a joint understanding-generation objective~\citep{Wu2024JanusDV,ma2025janusflow,chen2025blip3,chen2025janus,deng2025emerging} or prompt embeddings from Multimodal LLM (MLLM)-based encoders~\citep{wu2025qwen, fang2025got, wu2025omnigen2, wu2025openuni}. In earlier T2I models, spatial guidance in forms such as bounding boxes could be used to improve positional control~\citep{fang2025got, feng2023layoutgpt, yang2024mastering, lian2023llmgrounded, zhang2023controllablegpt4, zhang2024creatilayout} often via generating and stitching sub-images. In fact, stitching images dates back to classical methods~\citep{Davis1998MosaicsOS, Efros2001ImageQF}, though modern approaches use deep networks.  Methodologically, these externally position-guided modifications are related to tasks involving layout-conditioned T2I synthesis~\citep{zhang2023adding, Li2023GLIGENOG, farshad2023scenegenie, chen2024training, mou2023t2iadapter, ohanyan2024zeropainter, xie2023boxdiff, mo2024freecontrol}, 
which are more constrained by fine-grained user input. Unfortunately, these methods are implemented on older U-Net or Diffusion Transformer~\citep{peebles2023scalable} architectures. In contrast, many recent models adopt MMDiT~\citep{esser2024scaling} or related architectures~\citep{wu2025qwen, flux2024, cai2025hidream}, where such techniques often fail to generalize effectively~\citep{jiao2025unieditflow}. Instead, it has been shown that MMDiT-based architectures may be more amenable to generation-level edits like prediction swapping~\citep{jiao2025unieditflow, bader2025sub}, offering insights into MMDiT’s inference behavior that we build upon.

\paragraph{Position control benchmarks.} Many T2I benchmarks evaluate image generation capabilities, with GenEval~\citep{ghosh2023geneval} and T2I-CompBench++~\citep{huang2023t2icompbench, huang2025t2icompbench++} among the most widely used. Both include basic spatial reasoning categories of form $\langle obj_1\rangle \langle relation_{12}\rangle\langle obj_2\rangle$. %
While some benchmarks target positional understanding, many focus on basic formats lacking task complexity~\citep{Gokhale2022BenchmarkingSR}, or shift to related spatial tasks like shape generation~\citep{Sim2024EvaluatingTG} or 3D spatial reasoning~\citep{Wang2025GenSpaceBS}. Other general benchmarks include 2D spatial tasks~\citep{li2025unieval, wang2025lmm4lmm} which may extend to 3 or 4 objects~\citep{bakr2023hrs, feng2023layoutgpt}, or evaluate prompts that are \textit{natural}~\citep{feng2023layoutgpt} or long~\citep{hu2024ella}, but lack explicit structure or precise positional tasks. Our proposed \benchmarkname advances positional evaluation with a comprehensive, targeted suite of complex tasks.

\section{Methods}
Despite progress in many aspects of image synthesis, position-related prompts continue to pose challenges for T2I generation models. To address this, we introduce \methodname, a flexible method to improve positional generation in models with MMDiT~\citep{esser2024scaling} or similar architectures.
\subsection{Preliminaries}
Recent generative models define generation as a time-dependent transformation from Gaussian noise \(x_0 \sim \mathcal{N}(0,I)\) to data \(x_1 \sim p_{\text{data}}\), expressed as
$
x_\tau = \alpha_\tau x_0 + \sigma_\tau x_1,
$
with \(\alpha_\tau\) decreasing and \(\sigma_\tau\) increasing over time $\tau$. FM models~\citep{albergo2023building, liu2023flow, lipman2023flow, esser2024scaling} treat this trajectory as a differential equation, with learned approximation of its conditional vector field.  
Simulating this equation from noise to data in $T$ steps yields samples.

Many models operate in a VAE latent space~\citep{kingma2022auto}, encoding an image into latent token sequence \(\mathcal{Z}_v = \{ z_v^i \}_{i=1}^{N_v}\). To condition generation, a text encoder (e.g. T5~\citep{Raffel2019ExploringTL} or CLIP~\citep{Radford2021LearningTV}) embeds the prompt as \(\mathcal{Z}_t = \{ z_t^i \}_{i=1}^{N_t}\). Learnable projections \(E_v\) and \(E_t\) map both sets into a shared embedding space, yielding sets of \textit{visual and text vectors}:
\[
\mathcal{X}_v = \{ x_v^i \}_{i=1}^{N_v} = E_v(\mathcal{Z}_v), \quad
\mathcal{X}_t = \{ x_t^i \}_{i=1}^{N_t} = E_t(\mathcal{Z}_t),
\]
where each \(x_v^i, x_t^i \in \mathbb{R}^d\) is a single vector in the corresponding sequence. MMDiT-like~\citep{esser2024scaling} architectures process \(\mathcal{X}_v\) and \(\mathcal{X}_t\) iteratively with a stack of time-conditioned transformer blocks.
Each block first applies a pre-attention transformation
$
(\tilde{\mathcal{X}}_v, \tilde{\mathcal{X}}_t) = F_\text{pre}(\mathcal{X}_v, \mathcal{X}_t; \tau),
$
which includes normalization and timestep modulation. Let
$
\tilde{\mathcal{X}} = \tilde{\mathcal{X}}_v \cup \tilde{\mathcal{X}}_t = \{\tilde{x}_i\}_{i=1}^{N}, N = N_v + N_t
$
denote all pre-attention vectors. Each $\tilde{x}_i \in \tilde{\mathcal{X}}$ is mapped to queries, keys, and values via functions $Q, K, V$ derived from linear projections and normalization. Attention outputs are then computed as:
$$
z(\tilde{x}_i) = \sum_{\tilde{x}_j \in \tilde{\mathcal{X}}} \text{softmax}_j \Bigg( \frac{\langle Q(\tilde{x}_i), K(\tilde{x}_j) \rangle}{\sqrt{d}} + M(\tilde{x}_i, \tilde{x}_j) \Bigg) V(\tilde{x}_j),
$$
where $M : \tilde{\mathcal{X}} \times \tilde{\mathcal{X}} \to \{0, -\infty\}$ is an attention mask, and
$
\mathcal{Z} = \{ z(\tilde{x}_i) \}_{i=1}^{N}
$
denotes the set of attention outputs. Finally, a post-attention transformation
$
(\mathcal{X}_v^*, \mathcal{X}_t^*) = F_\text{post}(\mathcal{Z}, \mathcal{X}_v, \mathcal{X}_t; \tau)
$
updates the representations. Attention is often multi-headed, where each head $h$ computes attention independently over $\tilde{\mathcal{X}}$ with its own $Q_h, K_h, V_h$, and the outputs are combined into $\mathcal{Z}$. \methodname builds upon MMDiT's design, particularly with targeted masking within the multimodal transformer blocks.

\begin{figure}[t]
    \centering
    \includegraphics[width=1.0\textwidth]{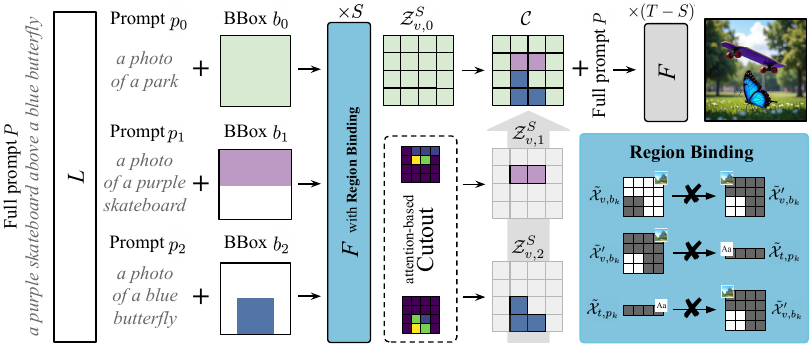} %
    \caption{\methodname: Multimodal LLM $L$ splits full prompt $P$ into object prompts ${p_k}$ and bounding boxes ${b_k}$, along with full-image background prompt $p_0$. MMDiT-based model $F$ seperately sketches objects and background (\textit{butterfly}~\textcolor{blue_stitch}{\rule{1ex}{1ex}}, \textit{skateboard}~\textcolor{purple_stitch}{\rule{1ex}{1ex}}, \textit{park}~\textcolor{green_stitch}{\rule{1ex}{1ex}}) for $S$ timesteps. With \phasetwo{Region Binding} attention-masking constraints, $F$ generates each $p_k$ in $b_k$. In \textbf{Cutout}, the highest attention weights in a targeted head select tighter latent regions linked to foreground objects $\mathcal{Z}_{v,k}^{S}$, which are merged with the background latents $\mathcal{Z}_{v,0}^{S}$ to form composite latent $\mathcal{C}$. For the remaining steps, the unconstrained $F$ seamlessly stitches the sketches into a coherent image conditioned on the full prompt.}
    \label{fig:method}
\end{figure}

\subsection{\methodname: Training-Free Position Control for MMDiT}

As seen in Figure~\ref{fig:method}, \methodname begins by using MLLM $L$ to decompose the full text prompt \(P\) into \(K\) sub-prompts \(\{p_k\}_{k=1}^K\), each associated with a corresponding bounding box \(\{b_k\}_{k=1}^K\), where $b_k = [x_{\min}, x_{\max}, y_{\min}, y_{\max}] \in \{0, \dots, W-1\}^4$.
Each pair \((p_k, b_k)\) represents a distinct object in the scene. Additionally, \(L\) generates a background prompt \(p_0\), with a bounding box \(b_0\) that spans the entire image. For the first $S$ steps, the FM model $F$ separately generates everything within the designated bounding boxes via Region Binding (detailed below). Once objects are adequately formed, their corresponding foreground latent tokens $\mathcal{Z}_{v,k}^{S}$ are cut from the resulting latent maps. We do so with our Cutout (discussed later in this section), the necessary information for which is obtained from a model-specific attention head. The extracted foreground latent tokens $\mathcal{Z}_{v,k}^{S}$ are combined with the common background latent tokens $\mathcal{Z}_{v,0}^{S}$ into a single composite latent $\mathcal{C}$, used exclusively for the remainder of the generation. The model proceeds without constraints and conditioned on full prompt $P$, allowing $F$ to enhance the overall quality and consistency while completing the image.

\paragraph{\phasetwo{Region Binding}}
\label{sec:phase1_constraints}

To ensure objects are fully generated within their specified bounding boxes and to prevent capturing of partial fragments during foreground extraction, we introduce three attention-masking constraints. These constraints guide the model to focus generation solely within each bounding box, effectively isolating the object from the surrounding context.

Formally, for each sub-prompt $p_k$ and bounding box $b_k$, let 
$\mathcal{\tilde{X}}_{v,b_k} \subseteq \mathcal{\tilde{X}}_{v}$ denote the visual vectors inside the bounding box, 
$\mathcal{\tilde{X}}_{v,b_k}' = \mathcal{\tilde{X}}_v \setminus \mathcal{\tilde{X}}_{v,b_k}$ those outside, and 
$\mathcal{\tilde{X}}_{t, p_k}$ the text vectors corresponding to the sub-prompt. 
In every layer and head we apply the attention mask $M$ with the following constraints: 
(1) block attention from inside to outside the bounding box, 
(2) block attention from outside the bounding box to the text, 
(3) block attention from the text to outside the bounding box:
\[
\begin{aligned}
M(\mathcal{\tilde{X}}_{v, b_k}, \mathcal{\tilde{X}}_{v, b_k}') = -\infty, \quad
M(\mathcal{\tilde{X}}_{v, b_k}', \mathcal{\tilde{X}}_{t, p_k}) = -\infty, \quad
M(\mathcal{\tilde{X}}_{t, p_k}, \mathcal{\tilde{X}}_{v, b_k}') = -\infty.
\end{aligned}
\]
The background bounding box $b_0$ spans the entire image, so $\mathcal{\tilde{X}}_{v, b_0} = \mathcal{\tilde{X}}_v$, imposing no constraints.
\paragraph{Cutout}
\label{sec:cutout}

To avoid visible seams from mismatched backgrounds, we extract the object latent tokens before constructing the composite latent $\mathcal{C}$. As generation is incomplete after step $S$, conventional segmentation tools like SAM~\citep{kirillov2023segment} cannot be used to find the shape to cut out. However, we observe that in some attention heads, text vectors focus on foreground objects. Given such a head (selection discussed in Section~\ref{sec:experiments_cutout}), we construct the Cutout mask by selecting visual tokens in descending order of attention weight. The attention weight for each visual token is computed as the average attention that its corresponding visual vector gets from all non-padding text vectors. %
We select tokens until their attention weights sum to a fraction $\eta \in [0,1]$ of the total attention assigned to all visual tokens. The mask is then smoothed with 2D max pooling of kernel size $\kappa$. 

\section{\benchmarkname: Benchmarking Positional Generation Capabilities}
As models improve on basic \textit{Position} tasks of form $\langle \mathit{obj}_1\rangle\langle \mathit{relation}_{12}\rangle\langle \mathit{obj}_2\rangle$, broader benchmarks are needed to more thoroughly evaluate T2I spatial capabilities. We introduce \benchmarkname, featuring tasks going beyond basic \textit{Position} in difficulty, yet maintaining clear and specific task definitions. Each task is designed to isolate and evaluate a specific aspect of T2I capabilities. To ensure ease of use, \benchmarkname builds upon GenEval, using Mask2Former-based~\citep{cheng2021mask2former} object detection and procedural relation verification extended to the new tasks. We reuse the same set of objects $\langle \mathit{obj}\rangle$ and relations $\langle \mathit{relation}\rangle$, and adopt their protocol of generating 100 prompts per category and four images per prompt. Figure~\ref{fig:qualitative} and the Appendix show example prompts. The new tasks include:

\textbf{Two Objects (2 Obj)}: For completeness, we inherit GenEval's \textit{Position} task, unchanged.

\textbf{Three Objects (3 Obj):} We extend the basic \textit{Position} task to three objects linked by two spatial relations. Objects $\langle\mathit{obj}_1\rangle, \langle\mathit{obj}_2\rangle, \langle\mathit{obj}_3\rangle$ are arranged consecutively in a chain wrapped to fit on a $2\times2$ grid, ensuring one horizontal and one vertical relation. For a single $\langle\mathit{relation}_{ij}\rangle$ between adjacent $\langle\mathit{obj}_i\rangle$ and $\langle\mathit{obj}_j\rangle$, we use $f(\langle\mathit{relation}_{ij}\rangle):=$ ``The $\langle \mathit{obj}_i\rangle$ is $\langle \mathit{relation}_{ij}\rangle$ the $\langle\mathit{obj}_j\rangle$'' or ``The $ \langle\mathit{obj}_j\rangle$ is $\langle\mathit{relation}_{ji}\rangle$ the $\langle\mathit{obj}_i\rangle$'', with equal probability. The task prompts are built in the form:  
``A photo of a $\langle\mathit{obj}_1\rangle$, a $\langle \mathit{obj}_2\rangle$, and a $\langle\mathit{obj}_3\rangle$. $f(\langle\mathit{relation}_{12}\rangle)$. $f(\langle\mathit{relation}_{23}\rangle)$.'' Afterwards, objects in the first sentence are shuffled, along with sentences 2 and 3. Objects are listed upfront to avoid ambiguity. The model is tested on correctly interpreting the relations, not the underlying grid.

\textbf{Four Objects (4 Obj):} We further extend to four objects connected by four spatial relations, otherwise using the same setup as in the \textit{3 Obj} task.  
The task prompts are built in the form:  
``A photo of a $\langle\mathit{obj}_1\rangle$, a $\langle\mathit{obj}_2\rangle$, a $\langle\mathit{obj}_3\rangle$, and a $\langle\mathit{obj}_4\rangle$. $f(\langle\mathit{relation}_{12}\rangle)$. $f(\langle\mathit{relation}_{23}\rangle)$. $f(\langle\mathit{relation}_{34}\rangle)$. $f(\langle\mathit{relation}_{41}\rangle)$.''  
Again, objects in sentence 1 and sentences 2--5 are randomly permuted, and the model is evaluated on correctly interpreting the specified spatial relations.

\textbf{Positional Attribute Binding (PAB):} In the form ``a $\langle\mathit{attr}_1\rangle\langle\mathit{obj}_1\rangle\langle\mathit{relation}_{12}\rangle$ a $\langle\mathit{attr}_2\rangle\langle \mathit{obj}_2\rangle$", PAB extends \textit{Attribute Binding} (``a $\langle \mathit{attr}_1\rangle\langle\mathit{obj}_1\rangle$ and a $\langle\mathit{attr}_2\rangle\langle\mathit{obj}_2\rangle$") with inter-object positions.

\textbf{Negative Relations (Neg):} To evaluate understanding of \textit{Negative Relations}, we use prompts of form, ``a photo of a $\langle \mathit{obj}_1\rangle$ and a $\langle \mathit{obj}_2\rangle$, a $\langle \mathit{obj}_1\rangle$ is not $\langle \mathit{relation}_{12}\rangle$ a $\langle \mathit{obj}_2\rangle$", with objects listed for clarity. It is evaluated that: (1) both objects are present, and (2) $\langle \mathit{obj}_2\rangle$ appears anywhere \textit{except} in the specified relation to $\langle \mathit{obj}_1\rangle$. Prompts are derived from GenEval's position prompts by replacing the relation with a negation of the opposite relation, keeping the same target image (e.g. ``a photo of a dog right of a teddy bear" $\rightarrow$ ``a photo of a dog and a teddy bear, a dog is not left of a teddy bear"). 

\textbf{Relative Relations (Rel):} We evaluate understanding of relations \textit{relative to other relations}. Prompts contain three objects and two spatial relations, the first in the form $\langle \mathit{relation}_{ij}\rangle$ and the second defined relative to the first. Relative relations $\langle \mathit{rel. relation} \rangle$ can be same or opposite, each comprising half the prompts. The \textit{same} relation is ``on the same side of", while the \textit{opposite} relations include ``on the other side of", ``on the opposite side of", and ``on the contrary side of", for diversity. Prompts take the form, ``a photo of a $\langle \mathit{obj}_1\rangle$ $\langle \mathit{relation}_{12}\rangle$ $\langle \mathit{obj}_2\rangle$, and a $\langle \mathit{obj}_3\rangle$ $\langle \mathit{rel. relation}\rangle$ the $\langle obj_2\rangle$ $\langle prep\rangle$ the $\langle \mathit{obj}_1\rangle$", with $\langle prep\rangle \in \{\text{``for'', ``as''}\}$ chosen to be grammatically correct.

\newcommand{\std}[1]{\scriptsize\textcolor{gray}{$\pm$#1}}

\begin{table}[t]\centering
\setlength{\tabcolsep}{2pt}
\caption{\benchmarkname reveals that leading T2I models still struggle with complex positional prompts. However, \methodname boosts positional generation on 3 different MMDiT-based  models, achieving SOTA.}
\begin{tabular}{lccccccc}
\toprule
\textbf{Model} & \textbf{2 Obj} & \textbf{3 Obj} & \textbf{4 Obj} & \textbf{Neg} & \textbf{Rel} & \textbf{PAB} & \textbf{Avg.$\uparrow$} \\
\midrule
{\small\textit{\color{gray} No layout guidance}}            \\
FLUX.1 [Dev]{\scriptsize~\citep{flux2024}}                & 0.22\hspace{19px}     & 0.06\std{0.01}     & 0.02\std{0.00}     & 0.62\std{0.01}     & 0.03\std{0.01} &     0.15\std{0.02}     & 0.18     \\
SD3 Medium{\scriptsize~\citep{esser2024scaling}}          & 0.34\hspace{19px}     & 0.05\std{0.01}     & 0.01\std{0.00}     & 0.62\std{0.02}     & 0.05\std{0.01} & 0.14\std{0.01}     & 0.20     \\
SD3.5 Large{\scriptsize~\citep{esser2024scaling}}         & 0.34\hspace{19px}     & 0.06\std{0.00}     & 0.02\std{0.01}     & 0.64\std{0.03}     & 0.06\std{0.02} & 0.16\std{0.01}     & 0.21     \\
HiDream-I1-Full{\scriptsize~\citep{cai2025hidream}}       & 0.60\hspace{19px}     & 0.19\std{0.01}     & 0.10\std{0.00}     & 0.66\std{0.00}     & 0.09\std{0.01} &          0.29\std{0.01}     & 0.32     \\
BAGEL{\scriptsize~\citep{deng2025emerging}}               & 0.64\hspace{19px}     & 0.23\std{0.01}     & 0.16\std{0.02}     & 0.73\std{0.01}     & 0.07\std{0.01} & 0.22\std{0.02}     & 0.34      \\
OpenUni-B-512{\scriptsize~\citep{wu2025openuni}}          & 0.77\hspace{19px}     & 0.09\std{0.01}     & 0.03\std{0.00}     & 0.56\std{0.01}     & 0.01\std{0.00} & 0.60\std{0.01}     & 0.34      \\
Janus-Pro{\scriptsize~\citep{chen2025janus}}              & 0.79\hspace{19px}     & 0.14\std{0.01}     &  0.04\std{0.01}    & 0.69\std{0.02}     & 0.11\std{0.00} & 0.46\std{0.03}     & 0.37      \\
BLIP3-o{\scriptsize~\citep{chen2025blip3}}                & \textbf{0.87}\std{0.01}         & 0.15\std{0.00}     & 0.02\std{0.00}     & 0.55\std{0.01}     & 0.08\std{0.01} & 0.63\std{0.01}     & 0.38          \\
Qwen-Image{\scriptsize~\citep{wu2025qwen}}                & 0.76\hspace{19px}     & 0.40\std{0.01}     & 0.21\std{0.02}     & 0.49\std{0.00}     & 0.10\std{0.01} & 0.61\std{0.03}     & 0.43      \\
\arrayrulecolor{gray}
\hline
\arrayrulecolor{black}
{\small\textit{\color{gray}Layout guidance, with training}} \\
GoT{\scriptsize~\citep{fang2025got}}                      & 0.34\hspace{19px}     & 0.02\std{0.00}     & 0.00\std{0.00}     & 0.52\std{0.01}     & 0.02\std{0.01} & 0.09\std{0.01}     & 0.17      \\
LayoutGPT{\scriptsize~\citep{feng2023layoutgpt}}                      & 0.41\std{0.04}     & 0.18\std{0.03}     & 0.10\std{0.02}     & 0.37\std{0.03}     & 0.25\std{0.02} & 0.07\std{0.02}     & 0.23      \\
\arrayrulecolor{gray}
\hline
\arrayrulecolor{black}
{\small\textit{\color{gray}Layout guidance, training-free}} \\
RPG{\scriptsize~\citep{yang2024mastering}}                      & 0.28\std{0.00}     & 0.07\std{0.01}     & 0.01\std{0.00}     & 0.65\std{0.00}     & 0.06\std{0.01}  & 0.14\std{0.01}     & 0.20      \\
LMD{\scriptsize~\citep{lian2023llmgrounded}}                      & 0.74\std{0.01}     & 0.43\std{0.02}     & 0.23\std{0.01}     & 0.54\std{0.01}     & 0.35\std{0.01} & 0.46\std{0.01}     & 0.46\vspace{5px} \\

\textbf{\methodnamebf (ours) + SD3.5 Large}                 &  0.53\std{0.02}    & 0.22\std{0.03}               & 0.12\std{0.01}               & 0.79\std{0.00}               & 0.27\std{0.02}           & 0.37\std{0.02}               & 0.38      \\
{\quad Gain over SD3.5 Large}                     & \textcolor{ForestGreen}{+0.19}   & \textcolor{ForestGreen}{+0.16} & \textcolor{ForestGreen}{+0.10}      & \textcolor{ForestGreen}{+0.15}    & \textcolor{ForestGreen}{+0.21}   
& \textcolor{ForestGreen}{+0.21} &\textcolor{ForestGreen}{+0.17} \\
\textbf{\methodname (ours) + FLUX.1 [Dev]}                     & 0.70\std{0.02}      & 0.44\std{0.01} &  0.38\std{0.01}  & 0.83\std{0.02}  & \textbf{0.48}\std{0.02}          & 0.44\std{0.00}       & 0.55          \\
{\quad Gain over FLUX.1 [Dev]}                     & \textcolor{ForestGreen}{+0.48}     & \textcolor{ForestGreen}{+0.38}   & \textcolor{ForestGreen}{+0.36} & 
\textcolor{ForestGreen}{+0.21}    & \textcolor{ForestGreen}{+0.45}   & \textcolor{ForestGreen}{+0.29} & \textcolor{ForestGreen}{+0.37} \\
\textbf{\methodname (ours) + Qwen-Image}                     & \textbf{0.87}\std{0.01}      & \textbf{0.67}\std{0.01}  & \textbf{0.61}\std{0.01}   & \textbf{0.93}\std{0.00}  & 0.43\std{0.01}          & \textbf{0.77}\std{0.00}       & \textbf{0.71}          \\
{\quad Gain over Qwen-Image}                     & \textcolor{ForestGreen}{+0.11}     & \textcolor{ForestGreen}{+0.27}   & \textcolor{ForestGreen}{+0.40}  & \textcolor{ForestGreen}{+0.44} & \textcolor{ForestGreen}{+0.33}   & \textcolor{ForestGreen}{+0.16}  & \textcolor{ForestGreen}{+0.28}  \\
\bottomrule
\end{tabular}\label{tab:poseval}
\end{table}

\section{Main Experiments}
We validate our three primary contributions: \methodname, Cutout, and \benchmarkname. %
In \methodname, $S = 10$ for FLUX and SD3.5 and $6$ for Qwen-Image. For all three models, we use $T=50$, $\kappa=5$, and bounding boxes $b_k$ are generated on a $W\times W = 32 \times 32$ grid with prompts $p_k$ from GPT-5~\citep{gpt5} and GPT-4~\citep{openai2024gpt4technicalreport}. $\eta$ is set to $0.95$ for SD3.5 and FLUX. Results are on 3 seeds. All other parameters are the same as the base models. Our code is attached in the Supplementary Materials. 

\subsection{Enhancing Positional Understanding with \methodname}
\benchmarkname highlights that while SOTA models handle basic \textit{Position} prompts well, they struggle with complexity, suggesting that the overall positional problem remains unsolved. Table~\ref{tab:poseval} presents \benchmarkname results, covering our five newly introduced tasks and GenEval's \textit{Position} (\textit{2 Obj}). While models like BLIP3-o~\citep{chen2025blip3} and JanusPro~\citep{chen2025janus} perform the basic task well ($87\%$ and $79\%$, respectively) their accuracy drops sharply on harder tasks. On \textit{Relative Relations}, they score only $8\%$ and $11\%$, and when scaling to \textit{Four Objects}, their accuracy falls to $2\%$ and $4\%$. \benchmarkname reveals greater shortcomings in top models than the basic \textit{Position} task suggests.

\methodname consistently enhances base models on \benchmarkname tasks, with up to $48$ percentage point improvement ($218\%$ relative increase) over FLUX on \textit{2 Obj} and $37$ percentage point increase on \benchmarkname overall ($206\%$ relative increase). This boost is also seen on \textit{Position} tasks in other benchmarks (T2ICompBench~\citep{huang2023t2icompbench} and HRS-Bench~\citep{bakr2023hrs} in the Appendix). \methodname $+$ Qwen-Image, ranks first on $5/6$ \benchmarkname tasks and ties with \methodname $+$ FLUX on the last. It has the highest overall accuracy, with a $54\%$ relative gain over the previous best, LMD~\citep{lian2023llmgrounded}.

For Qwen-Image, FLUX, and SD3.5, \methodname refines positional details without degrading visual quality, shown in Figure~\ref{fig:qualitative} on all \benchmarkname categories (SD3.5 in Appendix). Even scaling to four objects, it can integrate without awkward or unnatural seams. The boost observed in Table~\ref{tab:poseval} from applying \methodname can be partly attributed to reduced positional errors, such as incorrectly placing or omitting objects. These mistakes are seen when FLUX omits the oven in \textit{Four Object}, and Qwen-Image incorrectly places the cow left of the laptop in \textit{Negative}. %
Moreover, \methodname demonstrates strong comprehension of linguistically challenging prompts, like those in \textit{Negative} and \textit{Relative Relation}.

Figure~\ref{fig:qualitative2} further highlights \methodname's handling of complex scenes on FLUX. On the left, we show how it successfully interprets a prompt that combines all six \benchmarkname task categories. Despite the complexity (featuring four distinct objects, multiple attributes, and both negative and relative spatial relations), \methodname generates a coherent and semantically accurate image. In the center, we demonstrate that it enhances FLUX's ability to generate rare and challenging combinations, such as a dinosaur \textit{above} a rainbow. On the right, we show that \methodname's strong language understanding enables it to interpret even confusing text and translate it into coherent visual arrangements. %
\begin{figure}[t]
    \centering
    \includegraphics[width=1.0\textwidth]{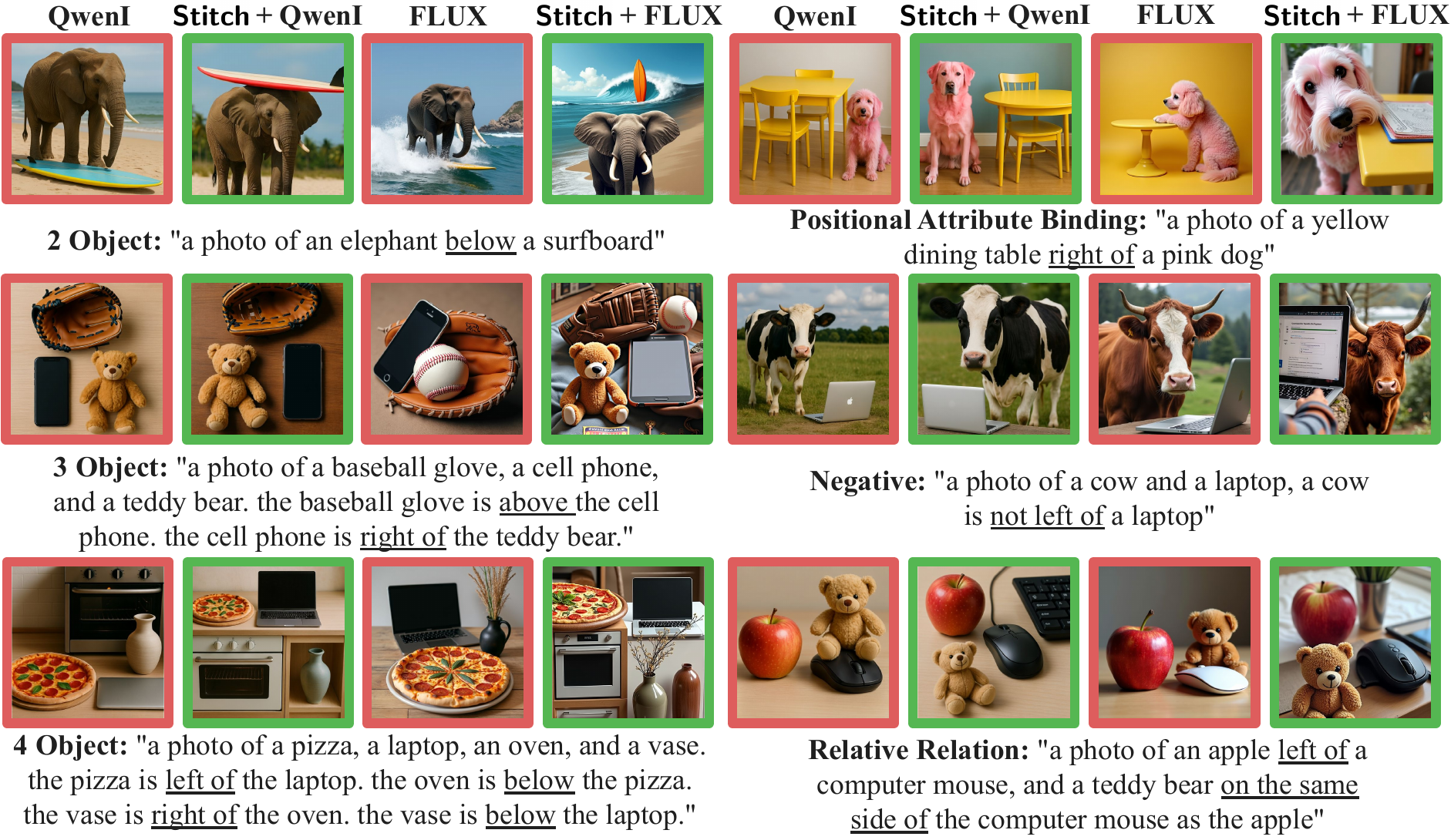} %
    \caption{\methodname corrects Qwen-Image (QwenI) and FLUX position on \benchmarkname without quality loss.}
    \label{fig:qualitative}
    \vspace{-16px}
\end{figure}
\subsection{Assessing Foreground Segmentation Accuracy of Cutout}
\label{sec:experiments_cutout}
We choose per-model attention heads by generating $80$ images with the prompt ``a photo of a $\langle \text{obj} \rangle$" using the $80$ GenEval objects. After step $S$, we save text-to-image attention averaged over non-padding tokens, avoiding unwanted behavior introduced by padding. We test thresholds $\eta \in \{0.75, 0.80, 0.85, 0.90, 0.95, 0.97, 0.99\}$ and extract SAM~\citep{kirillov2023segment} masks generated from the final images to use for ground truth, excluding objects with inadequate SAM masks. We evaluate the Intersection over Union (IoU) between masks generated by each attention head across the different $\eta$ values, choosing the head-threshold combination with the highest IoU. 

\begin{wrapfigure}{r}{0.31\textwidth}
    \vspace{-15px}
    \centering
\includegraphics[width=0.30\textwidth]{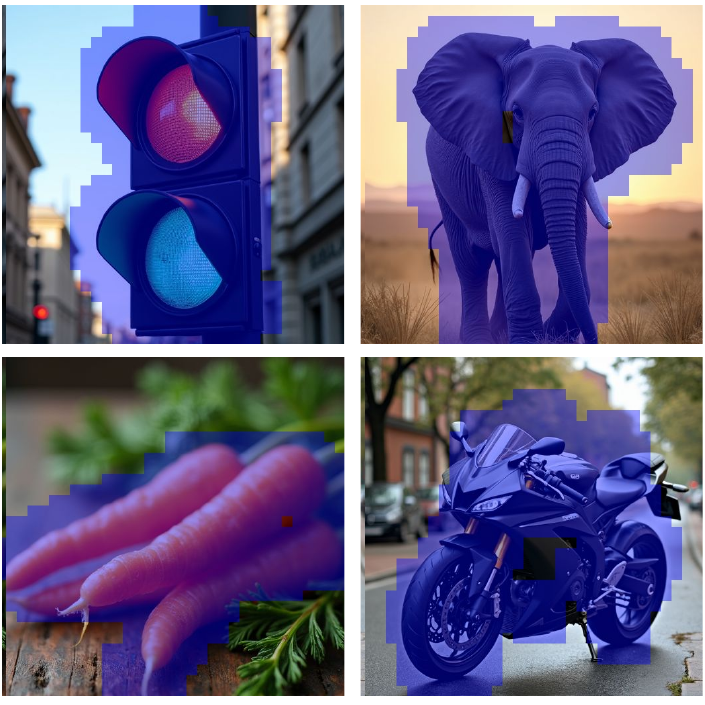} %
    \caption{Cutout cleanly extracts objects mid-generation.}
    \label{fig:seg_maps}
    \vspace{-14px}
\end{wrapfigure}

In the Appendix, we present the top five attention heads for each base model, ranked by IoU with SAM maps and displayed with each head's optimal threshold $\mu$. We also report Intersection over Target (IoT), calculated as $\frac{|\text{Prediction} \cap \text{Target}|}{|\text{Target}|}$. For each model, several attention heads exhibit strong IoU and IoT, indicating suitability for Cutout. For example, FLUX block 14's head 20 achieves IoU $=62\%$ and IoT $= 92\%$. Example masks are presented in Figure~\ref{fig:seg_maps} for this head, selected for FLUX Cutout. We observe that Cutout masks often leave a narrow border around the object, perhaps because low-level details have not yet solidified, helping explain their better IoT compared to IoU. This behavior suits \methodname's purpose: the border does not degrade the images and helps capture the full object, although even small missing parts can be reconstructed in later generation steps.

\subsection{Assuring Reliability of our \benchmarkname Benchmark}

\benchmarkname adopts the same evaluation protocol as GenEval~\citep{ghosh2023geneval}, whose components were thoroughly validated by the original authors. We conduct a user study to validate the reliability of the evaluation platform for the newly introduced benchmark categories. For this study, we create a candidate image pool with 4 images per prompt from Qwen-Image, \methodname + Qwen-Image, FLUX, \methodname + FLUX, SD3.5, and \methodname + SD3.5. We divide this pool into images labeled as \textit{correct} and \textit{incorrect} by the automated evaluation, then randomly select 50 images from each group. Three annotators independently evaluate whether each image ``correctly follows the prompt", and we report their level of agreement with the automatic evaluation. Figure~\ref{fig:poseval} shows the evaluation protocol performs consistently (table in Appendix), with all \benchmarkname categories within $10\%$ of the average inter-annotator agreement. 
\begin{wrapfigure}{r}{0.31\textwidth}
    \centering
    \vspace{-18px}
    \hspace{-10px}
    \includegraphics[width=0.33\textwidth]{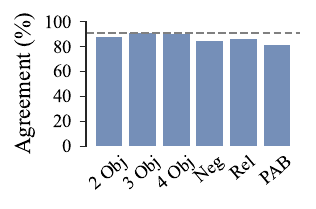} %
    \vspace{-20px}
    \caption{PosEval's evaluation is human-aligned, near average annotator agreement (line \textcolor{gray}{-\,-}).}
    \vspace{-10px}
    \label{fig:poseval}
\end{wrapfigure}

Each \benchmarkname task is challenging yet complementary, allowing insights through comparison. Comparing \textit{2 Obj}, \textit{3 Obj}, and \textit{4 Obj} reveals how models handle positional generation as object count and relationships grow. As shown in Table~\ref{tab:poseval}, strong \textit{2 Obj} performance does not guarantee scaling. BLIP3-o~\citep{chen2025blip3} scores $87\%$ on \textit{2 Obj} but drops to $15\%$ and $2\%$ on \textit{3 Obj} and \textit{4 Obj}, while Qwen-Image~\citep{wu2025qwen}, with lower initial scores, maintains better performance at $40\%$ and $21\%$, respectively. This suggests some models may overfit to previously seen tasks without effectively handling positional generation as complexity grows. Moreover, although \textit{Negative} prompts generate the same images as \textit{2 Obj} prompts, they are easier to satisfy due to looser criteria (e.g., “not right” vs. “left”). If a model scores higher on \textit{2 Obj} than \textit{Neg} (e.g. Janus-Pro’s $79\%$ vs. $69\%$), the gap likely reflects prompt interpretation issues rather than image generation. Similarly, both \textit{Rel} and \textit{3 Obj} include three objects and two spatial relations, but \textit{Rel} is linguistically more complex and restricts relations to one axis, while \textit{3 Obj} spans two axes. For example, Qwen-Image scores $40\%$ on \textit{3 Obj} but only $10\%$ on \textit{Rel}, and $76\%$ on \textit{2 Obj} versus $49\%$ on \textit{Neg}, indicating difficulty with complex prompts.

\section{Additional Analysis and Ablations of \methodname}
\label{sec:ablations}
\begin{wrapfigure}{r}{0.36\textwidth}
    \centering
    \vspace{-15px}    \includegraphics[width=0.36\textwidth]{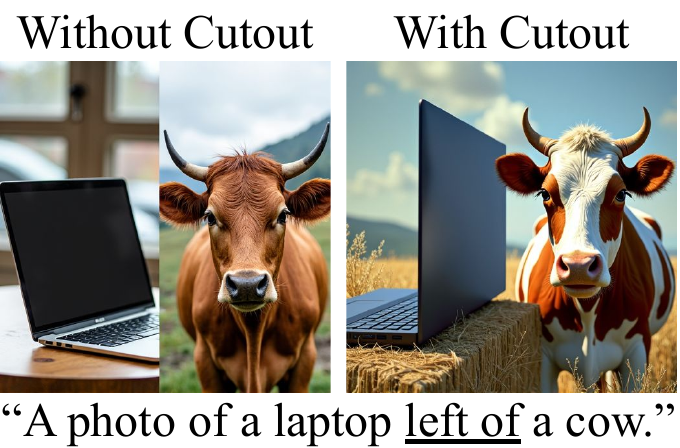} %
    \vspace{-15px}
    \caption{Cutout boosts coherence.}
    \label{fig:seg_ablation}
    \vspace{-40px}
\end{wrapfigure}

We investigate \methodname in greater detail. Table~\ref{tab:ablation}, presents an ablation study on FLUX to evaluate Region Binding and Cutout's contributions. After choosing a Cutout head, $\eta$ is raised to 0.95 to widen the object buffer. This choice is also ablated. We test accuracy on \benchmarkname and use a human study to evaluate \textit{Blend} on the basic GenEval \textit{Position} task, \textit{2 Obj.} We define an image having \textit{Blend} as being ``visually coherent (and not like many images put side by side)". Figure~\ref{fig:seg_ablation} illustrates this effect: the right image (with Cutout) exhibits \textit{Blend}, while the left image (without Cutout) does not. For each setup, three people independently and binarily decide if each of 100 images (one image per prompt) displays \textit{Blend}, and we report the percentage of images for which they answer \textit{yes}.

As shown in Table~\ref{tab:ablation}, the Region Binding is the primary driver of \methodname's position-related performance improvements (e.g. boosting \textit{2 Obj} from $22\%$ to $81\%$). However, combining full bounding boxes requires using backgrounds from the partially generated Region Binding predictions, which may contain conflicting color information; this can hinder the model's ability to seamlessly stitch components into a coherent image. Consequently, Region Binding alone fails to properly blend $32\%$ of images. Adding Cutout improves the \textit{Blend} success rate to $99\%$ of images, though at the cost of some positional accuracy. Raising the Cutout threshold to $\eta=0.95$ recovers lost performance. While lower thresholds often aid in selecting heads focused on the foreground, a higher threshold is preferable afterward, as \methodname benefits from higher IoT. As shown in Table~\ref{tab:ablation}, this improves accuracy without significantly affecting \textit{Blend}. This final combination is what allows \methodname to achieve both precise positional control and high image quality.

\begin{table}[t]
\centering
\setlength{\tabcolsep}{3pt}
\caption{While Region Binding (RB) drives \methodname's positional control, Cutout improves the blend. By increasing Cutout threshold $\eta$ to 0.95, we recover accuracy without sacrificing blend.}
\label{tab:ablation}
\begin{tabular}{cccccccccc}
\toprule
\phasetwo{RB} & \textbf{Cutout} &  $\eta=0.95$ & \textbf{2 Obj} & \textbf{3 Obj} & \textbf{4 Obj} & \textbf{Neg} & \textbf{Rel} & \textbf{PAB} & \textbf{Blend} \\
\midrule
 & & & 0.22 & 0.06\std{0.01} & 0.02\std{0.00} & 0.62\std{0.01} & 0.03\std{0.01} & 0.15\std{0.02} & 1.0 \\
\cmark & & & 0.81\std{0.02} & 0.58\std{0.03} & 0.52\std{0.02} & 0.90\std{0.01} & 0.63\std{0.01} & 0.51\std{0.01} & 0.68 \\
\cmark & \cmark & & 0.51\std{0.01} & 0.30\std{0.01} & 0.18\std{0.01} & 0.71\std{0.03} & 0.31\std{0.04} & 0.29\std{0.01} & 0.99 \\
\cmark & \cmark & \cmark & 0.70\std{0.02} & 0.44\std{0.01} & 0.38\std{0.01} & 0.83\std{0.02} & 0.48\std{0.02} & 0.44\std{0.00} & 0.95 \\
\bottomrule
\end{tabular}
\vspace{-10px}
\end{table}

Preserving base model quality while enhancing positional capabilities is essential. To validate, we compare each base model's quality with \methodname's via Aesthetic Score~\citep{aestheticscore}. Across four images per prompt on the six \benchmarkname tasks, \methodname results in only marginal changes: from $6.2\pm0.8$ to $6.1\pm0.8$ with Qwen, $6.3\pm0.8$ to $6.1\pm0.7$ with FLUX, and $5.3\pm0.8$ to $5.1\pm0.7$ with SD3.5. This indicates that \methodname does not significantly degrade image quality.

\begin{wrapfigure}{r}{0.36\textwidth}
    \centering
    \vspace{10px}
\includegraphics[width=0.36\textwidth]{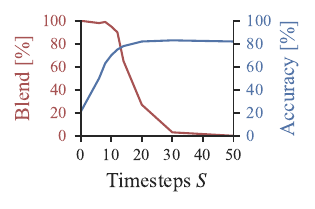}
    \vspace{-24px}
    \caption{In \methodname, increasing the Region Binding timesteps $S$ improves accuracy but reduces blend.}
    \label{fig:timestep_ablation}
    \vspace{-16px}
\end{wrapfigure}
To ensure diversity is preserved, we compute per-prompt mean pairwise distance between samples in DINOv2~\citep{oquab2024dinov} embedding space. 
We find average diversity is preserved or improved: FLUX increases from $0.34 \pm 0.15$ to $0.38 \pm 0.16$; SD3.5 remains at $0.43 \pm 0.15$ from $0.43 \pm 0.14$; and Qwen-Image rises from $0.16 \pm 0.11$ to $0.22 \pm 0.10$.

Finally, we investigate timestep selection $S$ for Region Binding on FLUX. In Figure~\ref{fig:timestep_ablation}, we vary the number of timesteps $S$ on the basic \textit{Position} task and compare to \textit{Blend}. Selecting later steps improves \textit{Position} accuracy but reduces \textit{Blend}, with a steep decline occurring between 10 and 20 steps. Offering the best balance, we select timestep 10 for FLUX. Similarly, we choose step 10 for SD3.5 and 6 for Qwen-Image.

\section{Conclusion}
We introduce \methodname, a training-free method for enhancing position-related T2I generation in MMDiT-based models. \methodname generates coherent, position-accurate images by first generating individual objects constrained to LLM-generated bounding boxes with our novel Region Binding constraints. With our Cutout, it then employs a targeted attention head to extract and combine these objects, before completing the image with the full prompt, unconstrained. We evaluate \methodname on \benchmarkname, our new benchmark extending GenEval with five challenging, position-focused tasks. Evaluating top models on \benchmarkname reveals that even the strongest T2I models still struggle with position-related tasks, despite performing well on basic \textit{Position} tasks. Yet \methodname significantly boosts Qwen-Image, FLUX, SD3.5, achieving SOTA results while remaining entirely training-free.

\section*{Reproducibility Statement}
We ensure the reproducibility of our experiments by providing detailed descriptions in Sections 5, 6, and the Appendix. The Supplementary Materials contain the implementation code, the prompts used in our benchmark, the bounding boxes generated by our method, as well as a README.md file with detailed reproduction instructions.

\section*{Acknowledgements}
This work was partially funded by the ERC (853489 - DEXIM) and the Alfried Krupp von Bohlen und Halbach Foundation, which we thank for their generous support. We are also grateful for partial support from the Pioneer Centre for AI, DNRF grant number P1. Mateusz Pach would like to thank the European Laboratory for Learning and Intelligent Systems (ELLIS) PhD program for support. The authors gratefully acknowledge the scientific support and resources of the AI service infrastructure \textit{LRZ AI Systems} provided by the Leibniz Supercomputing Centre (LRZ) of the Bavarian Academy of Sciences and Humanities (BAdW), funded by Bayerisches Staatsministerium für Wissenschaft und Kunst (StMWK). We would like to thank Jakub Pach for his assistance in conducting the user study.

\bibliography{iclr2026_conference}
\bibliographystyle{iclr2026_conference}

\clearpage
\appendix
\vspace{1cm}
\section*{Contents of Appendix}
\startcontents[sections]
\printcontents[sections]{}{1}{}

\clearpage
\section{Use of LLMs}
This work used LLMs to improve sentence clarity and flow.

\section{\methodname Results on Existing Benchmarks}
\begin{table}[h]\centering
\setlength{\tabcolsep}{2pt}
\caption{Results on tasks from existing benchmarks}
\begin{tabular}{lcccccc}
\toprule
\multirow{2}{*}{\textbf{Model}} 
  & \multicolumn{1}{c}{\textbf{GenEval}} 
  & \multicolumn{1}{c}{\textbf{T2ICompBench}} 
  & \multicolumn{3}{c}{\textbf{HRS-Bench (spatial)}} \\
& \multicolumn{1}{c}{Position} 
  & \multicolumn{1}{c}{Spatial} 
  & \multicolumn{1}{c}{Easy}
  & \multicolumn{1}{c}{Medium} 
  & \multicolumn{1}{c}{Hard}\\
\midrule
\textbf{SD3.5 Large}                          & 0.34 & 0.22 & 0.46 & 0.19 & 0.08  \\
\textbf{\methodname (ours) + SD3.5 Large}     & 0.53 & 0.28 & 0.54 & 0.26 & 0.08  \\
\quad Gain over SD3.5 Large                   & \textcolor{ForestGreen}{+0.19} & \textcolor{ForestGreen}{+0.06} & \textcolor{ForestGreen}{+0.08} & \textcolor{ForestGreen}{+0.07} & \textcolor{ForestGreen}{+0.00}\\
\textbf{FLUX.1 [Dev]}                         & 0.22 & 0.25 & 0.50 & 0.23 & 0.08 \\
\textbf{\methodname (ours) + FLUX.1 [Dev]}    & 0.70 & 0.44 & 0.82 & 0.52 & 0.23  \\
\quad Gain over FLUX.1 [Dev]                   & \textcolor{ForestGreen}{+0.48} & \textcolor{ForestGreen}{+0.19} & \textcolor{ForestGreen}{+0.32} & \textcolor{ForestGreen}{+0.29} & \textcolor{ForestGreen}{+0.15}\\
\textbf{Qwen-Image}                           & 0.76 & 0.36 & 0.73 & 0.45 & 0.25  \\
\textbf{\methodname (ours) + Qwen-Image}      & 0.85 & 0.50 & 0.88 & 0.63 & 0.40  \\
\quad Gain over Qwen-Image                   & \textcolor{ForestGreen}{+0.09} & \textcolor{ForestGreen}{+0.14} & \textcolor{ForestGreen}{+0.15} & \textcolor{ForestGreen}{+0.18} & \textcolor{ForestGreen}{+0.15}\\
\bottomrule
\end{tabular}\label{tab:geneval}
\end{table}

We present results on several \textit{Position}-related tasks in existing benchmarks: GenEval's \textit{Position} task~\citep{ghosh2023geneval}, T2I CompBench's \textit{Spatial} task~\citep{huang2023t2icompbench} (referred to as \textit{2D-Spatial} in T2ICompBench++~\citep{huang2025t2icompbench++}), and HRS-Bench's \textit{Spatial} task~\citep{bakr2023hrs}. T2I CompBench's \textit{Spatial} task includes two objects and a single relation. HRS Bench's \textit{Spatial} task is divided into three sub-categories: \textit{easy} prompts include two objects and one relation; \textit{medium} prompts include three objects and one or two relations; and \textit{hard} prompts include four objects and one or two relations.

With \methodname, we improve all three base models across all tasks, except HRS-Bench \textit{Hard} with SD3.5, where performance remains stable. This indicates that \methodname consistently enhances positional information across \textit{Position}-related tasks from multiple benchmarks.

\section{\benchmarkname User Study Results}
We include the numerical results of our user study verifying \benchmarkname's automated evaluation in Table~\ref{tab:user_study_benchmark_results}. As mentioned in the main paper, all results are within $10\%$ of the inter-annotator alignment ($91\%$).

\begin{table}[h!]
\centering
\caption{\benchmarkname alignment with human annotators. Average inter-annotator alignment is 91\%.}
\label{tab:user_study_benchmark_results}
\begin{tabular}{cccccc}
\toprule
\textbf{2 Obj} & \textbf{3 Obj} & \textbf{4 Obj} & \textbf{Neg} & \textbf{Rel} & \textbf{PAB} \\
\midrule
87.7\% & 90.7\% & 89.7\% & 84.7\% & 86.3\% & 81.3\% \\
\bottomrule
\end{tabular}
\end{table}

\clearpage
\section{Additional Qualitative Examples}
\begin{figure}[h!]
    \centering
    \includegraphics[width=\textwidth]{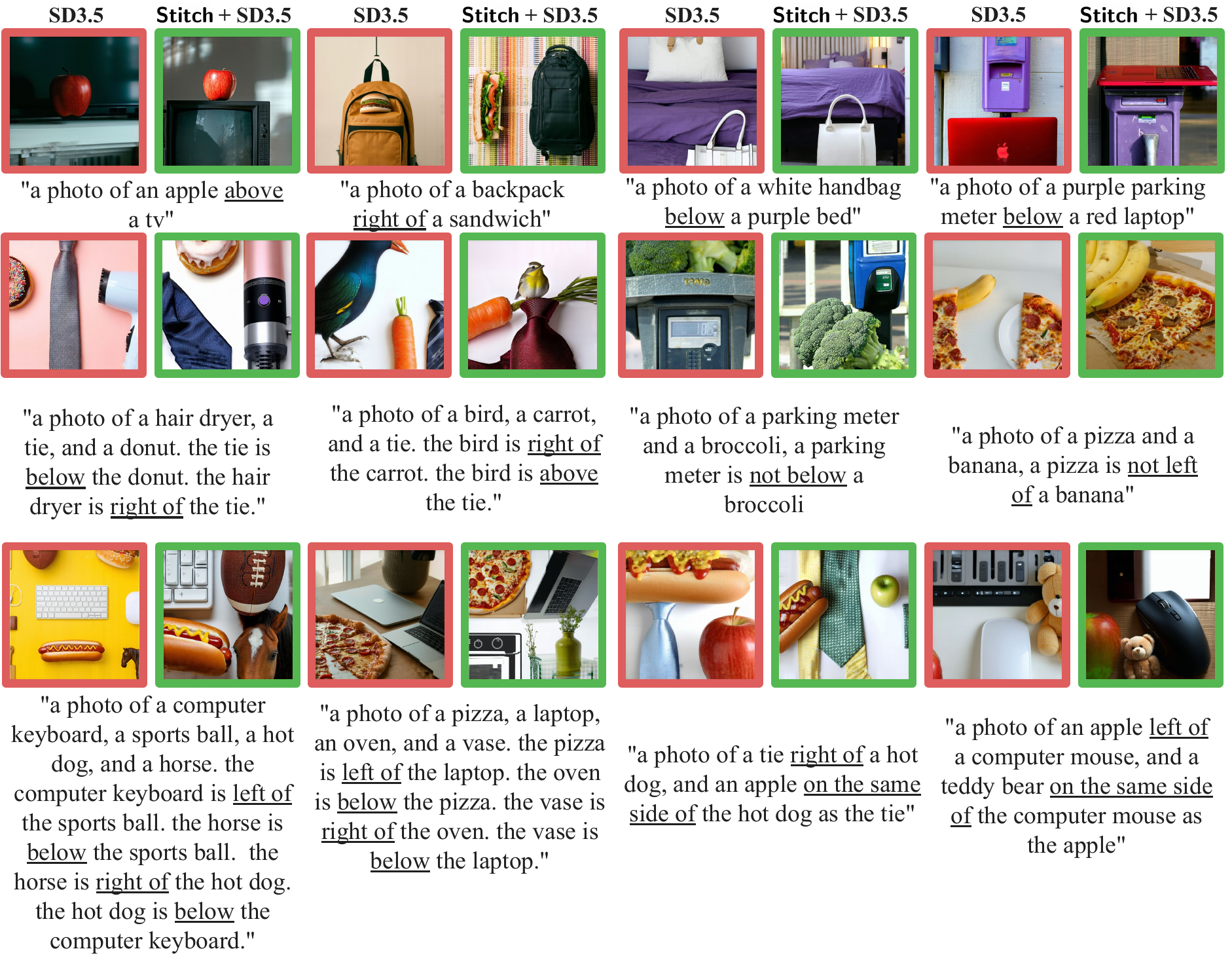} %
    \caption{Qualitative examples for \methodname + SD3.5.}
    \label{fig:qualitative-sd3}
\end{figure}

\begin{figure}[h!]
    \centering
    \includegraphics[width=\textwidth]{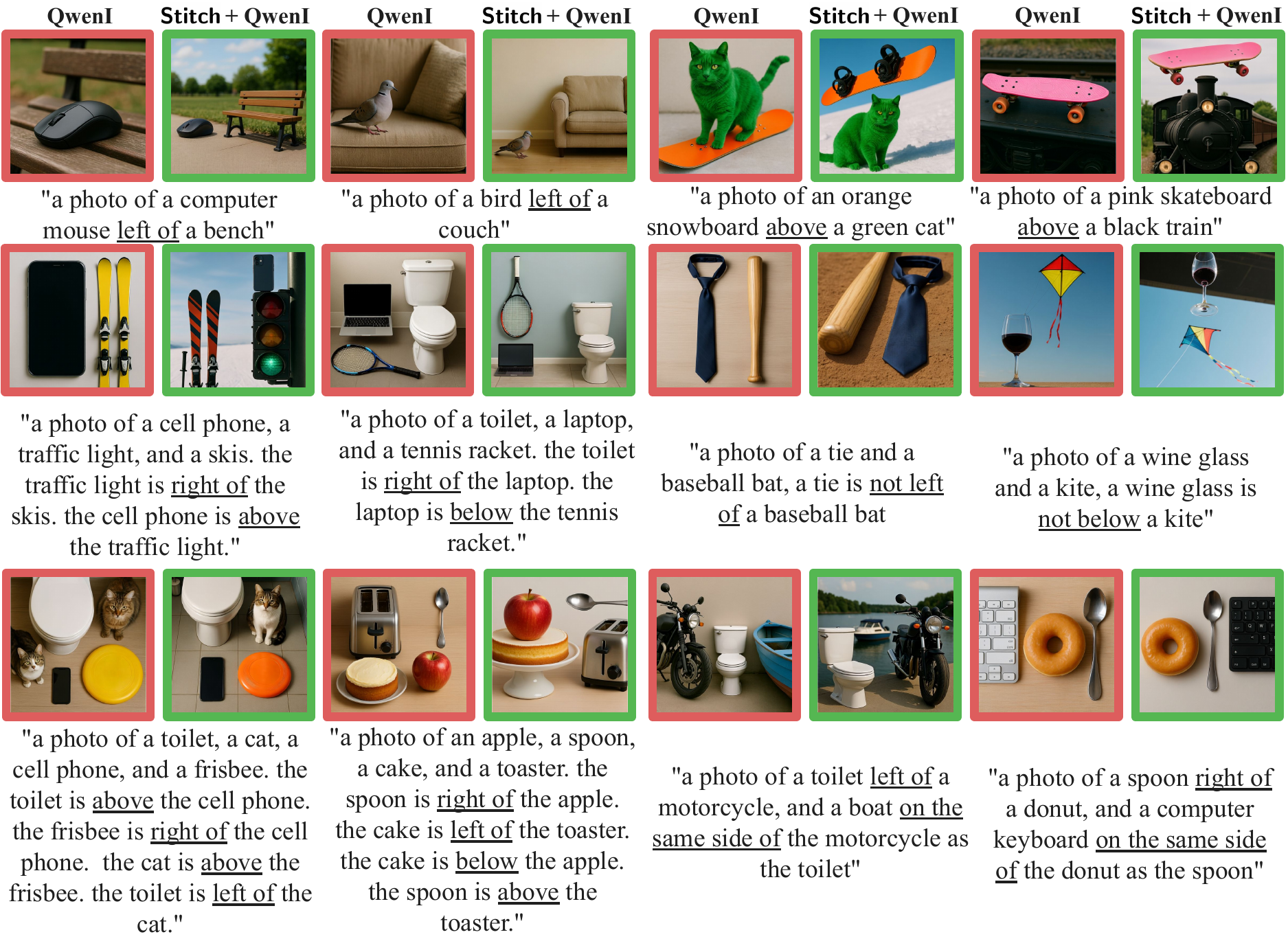} %
    \caption{Additional qualitative examples for \methodname + Qwen-Image (QwenI).}
    \label{fig:qualitative-qwen}
\end{figure}

\begin{figure}[h!]
    \centering
    \includegraphics[width=\textwidth]{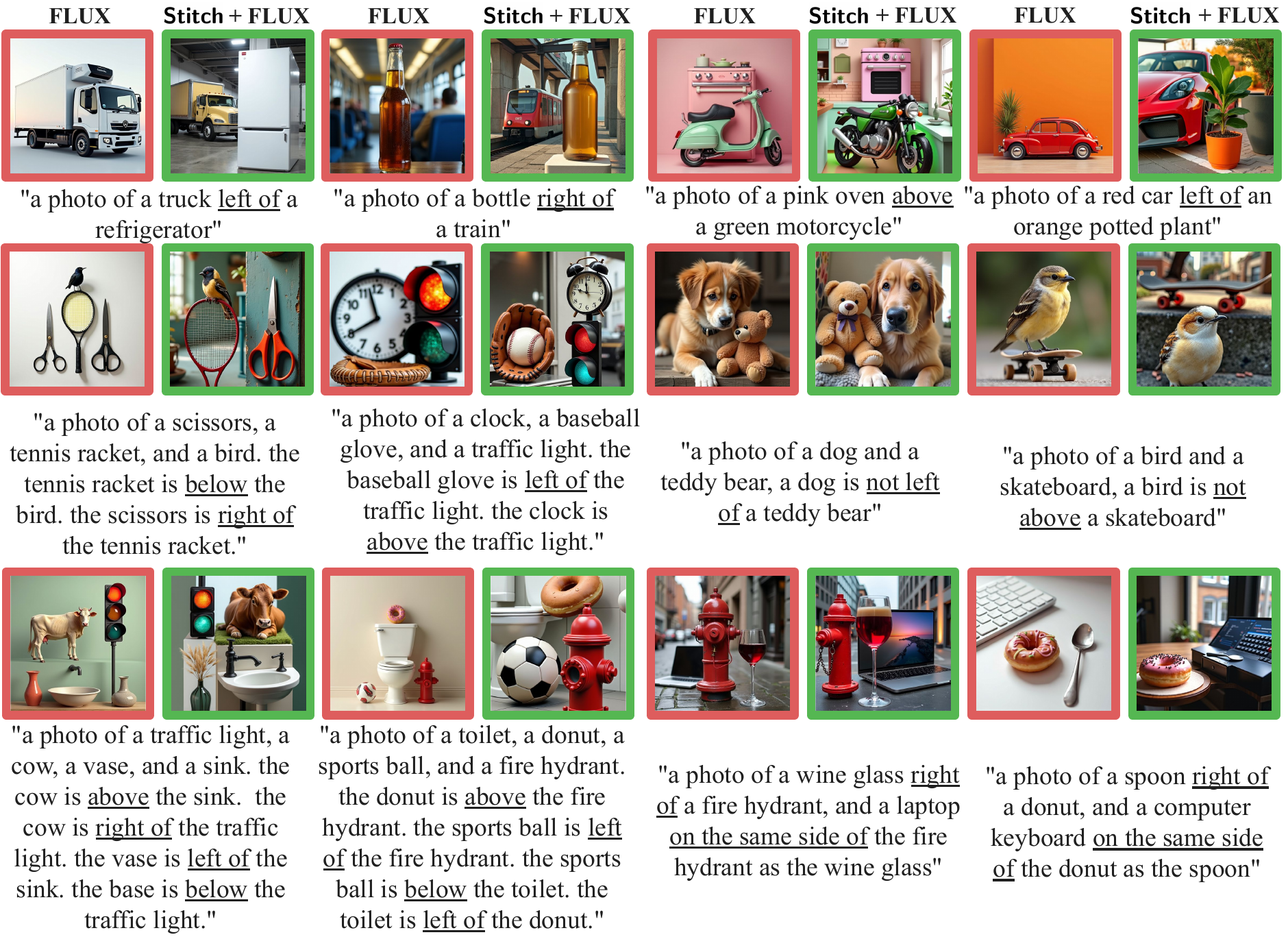} %
    \caption{Additional qualitative examples for \methodname + FLUX.}
    \label{fig:qualitative-flux}
\end{figure}

Figure~\ref{fig:qualitative-sd3} presents qualitative examples for \methodname combined with SD3.5, showcasing two examples from each \benchmarkname benchmark task. Additional qualitative results are provided for \methodname with FLUX in Figure~\ref{fig:qualitative-flux}, and for \methodname with Qwen-Image in Figure~\ref{fig:qualitative-qwen}, also on two prompts per task.

\clearpage

\section{Comparison of \benchmarkname with other Benchmarks}
We include an explicit comparison between \benchmarkname and previously existing benchmark tasks which include position-based information in Table~\ref{tab:benchmark_comparison}. We highlight several key aspects, focused on what is included in the benchmark and how specifically it is evaluated. First, we look at the number of objects present in positional tasks, starting with if they distinguish by the number of objects (\textit{Dist. \# Obj.}). For some benchmarks, such as the NSR-1k natural task~\citep{feng2023layoutgpt}, benchmark settings are not explicit about the number of objects present in each image. While all included benchmarks include 2-object prompts, we also specify if they include 3-object (\textit{3 Obj.}) and 4-object (\textit{4 Obj.}) prompts. Similar to \textit{Dist. \# Obj.}, we indicate if the tasks are generally focused (\textit{Focused tasks}), meaning that they include settings which separate the positional problem into specific sub-tasks. An example of non-focused tasks is DPG~\citep{hu2024ella} which targets the understanding of long prompts, but as a result does not sub-divide if prompts also test attribute binding, what types of positions are used (e.g. 2D vs. 3D), etc. In \textit{Attr.}, we indicate if attributes (e.g. color) are present. Finally, \textit{Neg. Pos.} shows the presence of \textit{negative} in the sense of \textit{position} (e.g. being "not left").

First, we observe that no existing benchmark incorporates the concept of \textit{negative position}. Additionally, none of the prior benchmarks simultaneously include both attributes and focused tasks. However, focused tasks are essential for thoroughly evaluating and advancing positional generation. They enable developers to pinpoint where models fail and clearly compare capabilities. This is a key advantage of our \benchmarkname, which addresses a gap left by previous benchmarks. In particular, our \textit{Positional Attribute Binding} task introduces a uniquely challenging and previously unexplored dimension. Our \textit{Relative Relation} task is also the first to specifically explore these types of more challenging \textit{relative} relations, in an isolated and focused way. To ensure consistency with prior work, we also scale the number of objects to four. Overall, \benchmarkname fills a critical gap in the literature by offering focused positional tasks while appropriately scaling the difficulty.

\begin{table}[h!]
\centering
\setlength{\tabcolsep}{2pt}
\caption{\benchmarkname comparison with other existing benchmarks}
\label{tab:benchmark_comparison}
\begin{tabular}{lcccccccc}
\toprule
 & Dist. \# Obj. & 3 Obj. & 4 Obj. & Focused tasks & Attr. & Neg. Pos. \\
\midrule
HRS-Bench~\citep{bakr2023hrs} & \cmark & \cmark & \cmark & \cmark & \xmark & \xmark \\
NSR-1K - template~\citep{feng2023layoutgpt}  & \cmark & \xmark & \xmark & \cmark & \xmark & \xmark \\
NSR-1K - natural~\citep{feng2023layoutgpt}  & \xmark & \cmark & \cmark & \xmark & \cmark & \xmark \\
VISOR~\citep{Gokhale2022BenchmarkingSR} & \cmark & \xmark & \xmark & \cmark & \xmark & \xmark \\
DPG~\citep{hu2024ella} & \xmark & \cmark & \cmark & \xmark & \cmark & \xmark \\
T2I CompBench++~\citep{huang2025t2icompbench++} & \cmark & \xmark & \xmark & \cmark & \xmark & \xmark \\
GenEval~\citep{ghosh2023geneval} & \cmark & \xmark & \xmark & \cmark & \xmark & \xmark \\
\textbf{\benchmarkname (ours)} & \cmark & \cmark & \cmark & \cmark & \cmark & \cmark \\
\bottomrule
\end{tabular}
\end{table}

\clearpage
\section{Top Cutout Segmentation Heads for each Model}
We present the results of the top attention heads for FLUX (Table~\ref{tab:seg_heads_flux}), Qwen-Image (Table~\ref{tab:seg_heads_qwen}), and SD3.5 (Table~\ref{tab:seg_heads_sd}). The bold results are those selected for Cutout for each model. This is block 14 head 20 for FLUX, block 14 head 34 for SD3.5, and block 25 head 1 for Qwen-Image. Interestingly, we find that the best segmentation heads typically fall in the middle blocks (between 10 and 20 for FLUX and SD3.5, and between 15 and 30 for Qwen-Image, which has more blocks).

\begin{figure}[t]
    \centering
    \includegraphics[width=1\textwidth]{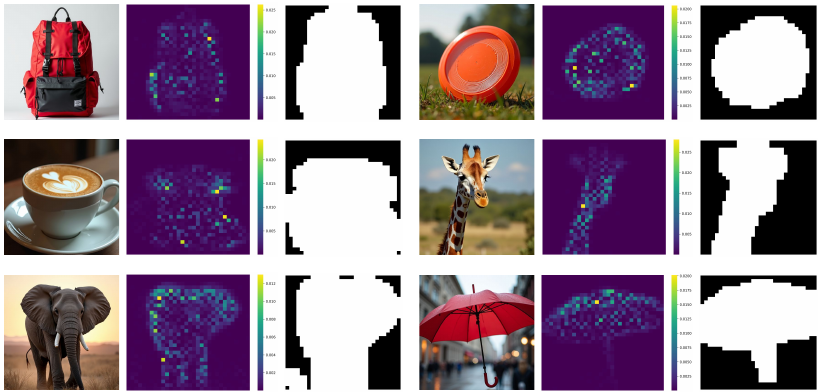} %
    \caption{Attention weights and Cutout masks from head 20 in block 14 of FLUX, at step 10 of 50.}
    \label{fig:attention_maps_flux}
\end{figure}

We include additional visualizations of the Cutout masks in FLUX in Figure~\ref{fig:attention_maps_flux}. For six example objects, we present: (1) the final image, (2) the attention map from the Cutout head (block 14, head 20 in FLUX) at step $S = 10$, and (3) the final Cutout masks thresholded at $95\%$. By comparing the attention maps to the final images, we observe that the selected head already encodes information about the object’s location in the latent space, even in the early stages of generation. Furthermore, the Cutout masks accurately capture the object's position, with a slight margin from the $95\%$ threshold. This buffer proves beneficial to \methodname, as confirmed by our ablation study.

\begin{figure}[t]
    \centering
    \includegraphics[width=0.3\textwidth]{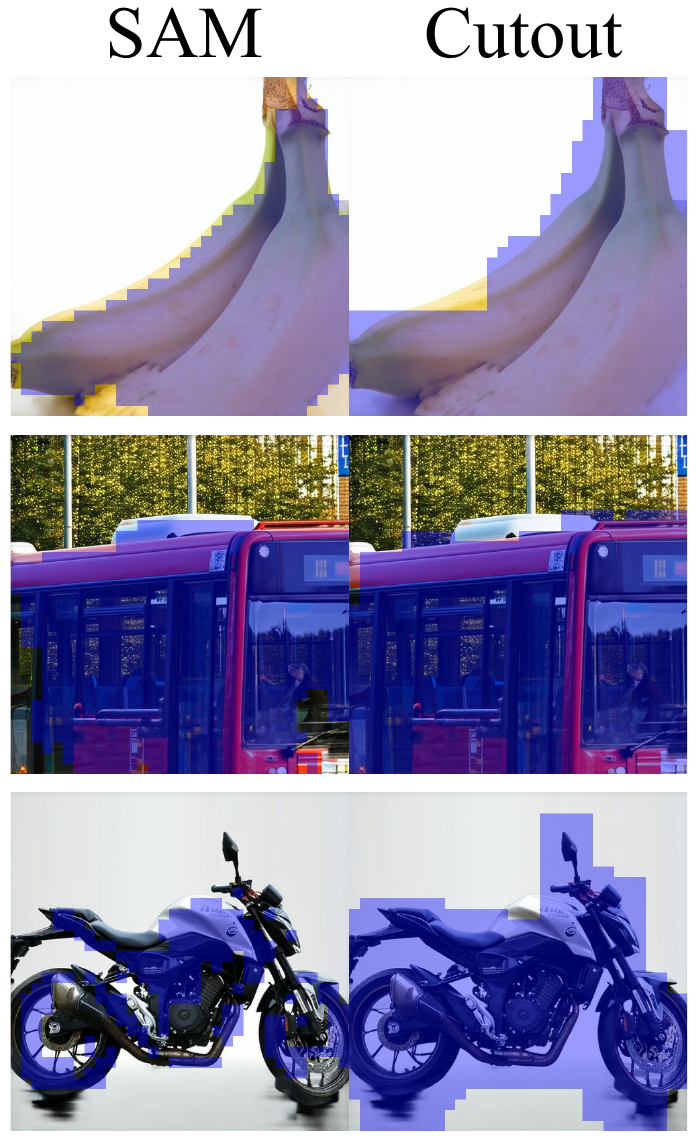} %
    \caption{Segmentation head maps for SD3.5.}
    \label{fig:seg_maps_sd3}
\end{figure}

In Figure~\ref{fig:seg_maps_sd3}, we also include segmentation maps for SD3.5, similar to those provided for FLUX in the main paper. These are compared to the SAM maps, for reference.

\begin{table}[h!]
\centering
\caption{Best segmentation heads for FLUX.}
\label{tab:seg_heads_flux}
\begin{tabular}{ccc|cc}
\toprule
Block & Head & $\eta$ & IoU & IoT \\
\midrule
\textbf{14} & \textbf{20} & 0.75 & \textbf{0.62} & \textbf{0.92} \\
17 & 23 & 0.75 & 0.56 & 0.82 \\
13 & 21 & 0.75 & 0.54 & 0.87 \\
14 & 15 & 0.75 & 0.54 & 0.87 \\
10 & 1 & 0.75 & 0.53 & 0.86 \\
\bottomrule
\end{tabular}
\end{table}

\begin{table}[h!]
\centering
\caption{Best segmentation heads for SD3.5.}
\label{tab:seg_heads_sd}
\begin{tabular}{ccc|cc}
\toprule
Block & Head & $\eta$ & IOU & IOT \\
\midrule
\textbf{14} & \textbf{34} & 0.75 & \textbf{0.56} & \textbf{0.94} \\
17 & 13 & 0.75 & 0.56 & 0.96 \\
15 & 31 & 0.75 & 0.55 & 0.94 \\
18 & 36 & 0.75 & 0.55 & 0.96 \\
14 & 6 & 0.80 & 0.55 & 0.90 \\
\bottomrule
\end{tabular}
\end{table}

\begin{table}[h!]
\centering
\caption{Best segmentation heads for Qwen-Image.}
\label{tab:seg_heads_qwen}
\begin{tabular}{ccc|cc}
\toprule
Block & Head & $\eta$ & IOU & IOT \\
\midrule
\textbf{25} & \textbf{1} & \textbf{0.9} & \textbf{0.55} & \textbf{0.86} \\
27 & 17 & 0.97 & 0.53 & 0.89 \\
26 & 4 & 0.85 & 0.53 & 0.82 \\
19 & 12 & 0.97 & 0.53 & 0.90 \\
18 & 11 & 0.85 & 0.52 & 0.84 \\
\bottomrule
\end{tabular}
\end{table}

\clearpage
\section{Bounding Boxes and Sub-prompts Generation}
To generate object bounding boxes $b_k$ and prompts $p_k$, $\forall k>0$, we utilize GPT-5~\citep{gpt5} model with reasoning effort set to minimal and system prompt:
\begin{verbatim}
You have a canvas of size {W}x{W}. Decompose the given description 
into single objects. Do not merge multiple objects into one box. 
Fully cover the canvas with the object boxes. Avoid overlaps. 
Do not leave space for background. Write 3 sentence justification 
and then output just valid JSON in the format: 
[
    {"prompt": "<object with properties only mentioned in the 
    description>", "x_min": , "y_min": , "x_max": , "y_max": }, 
    ...
]
\end{verbatim}
Each prompt $P$ is processed with user prompt:
\begin{verbatim}
Description: {P}
\end{verbatim}

To generate background prompt $p_0$ we use GPT-4~\citep{openai2024gpt4technicalreport} with system prompt:
\begin{verbatim}
Provide a simple fitting background description for this scene. 
The background must not mention any of the specific objects or 
elements from the prompt. The background must not mention any
other specific objects or people. Return only ONE word of the 
background text, nothing else.  
\end{verbatim}
Given a prompt $P$ the user prompt again takes form:
\begin{verbatim}
Description: {P}
\end{verbatim}

\clearpage
\section{User Study Interface}
\begin{figure}[h]
    \centering
    \includegraphics[width=\textwidth]
    {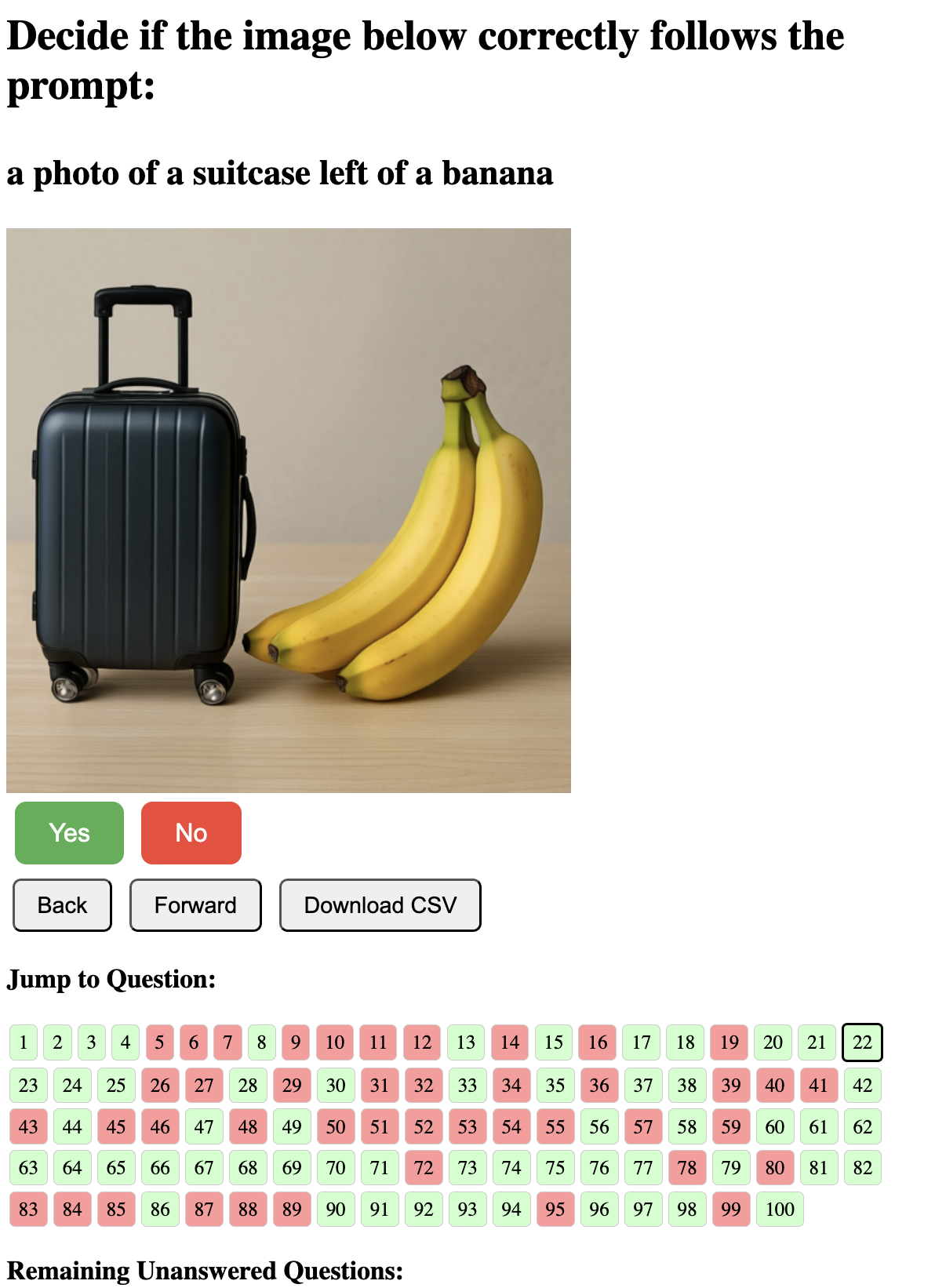} %
    \caption{\textit{PosEval} user study.}
    \label{fig:userstudy_poseval}
\end{figure}

\begin{figure}[h]
    \centering
    \includegraphics[width=\textwidth]
    {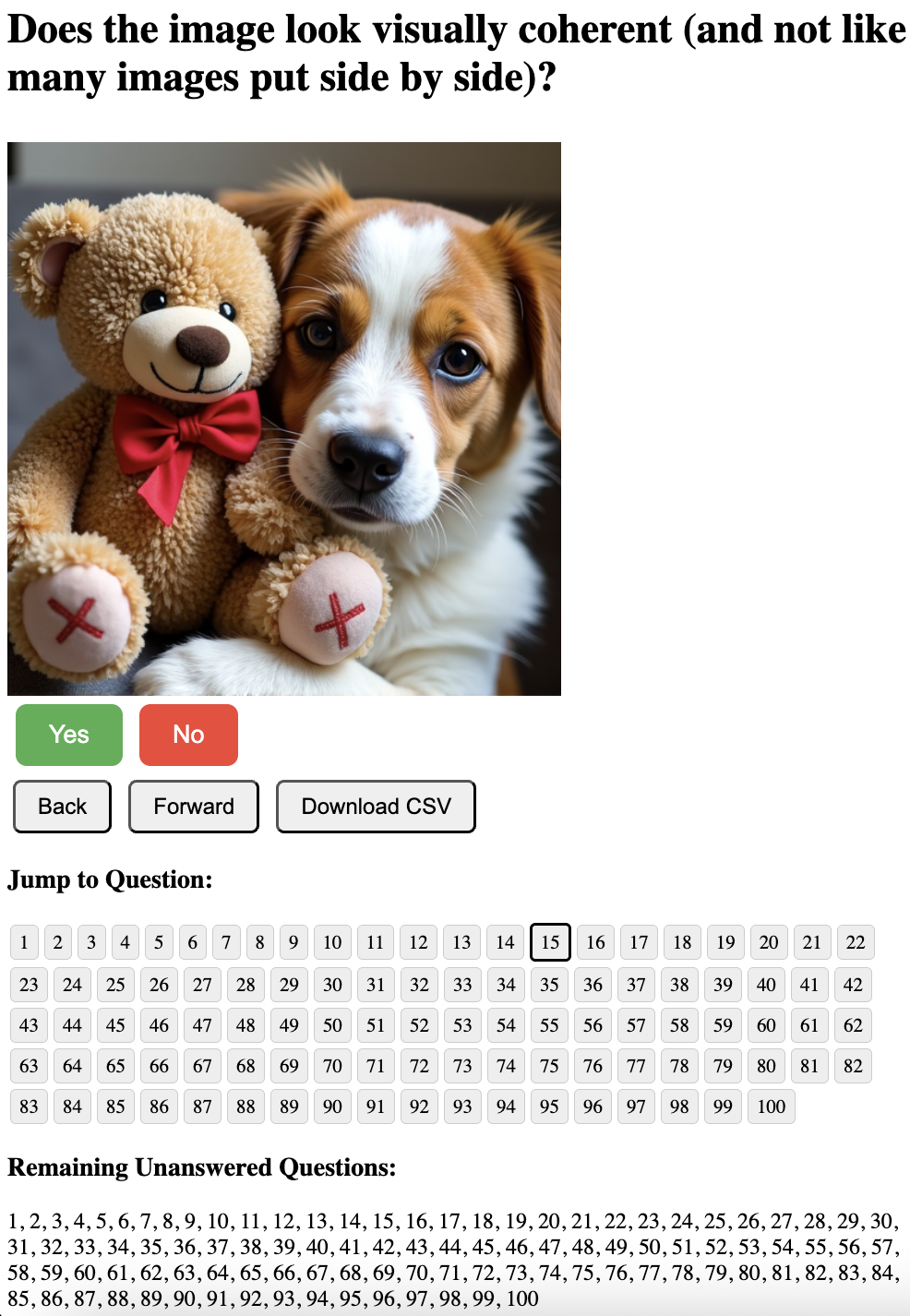} %
    \caption{\textit{Blend} user study.}
    \label{fig:userstudy_blend}
\end{figure}

In Figures~\ref{fig:userstudy_poseval} and~\ref{fig:userstudy_blend}, we include and example of the user interface for our user studies to measure \benchmarkname's quality and \textit{Blend} of the images.

\stopcontents[sections]

\end{document}